%% file: atar_mm26_camera_ready.tex
\PassOptionsToPackage{table}{xcolor}
\documentclass[sigconf]{acmart}
\makeatletter
\@ACM@balancefalse
\makeatother

\usepackage{multirow}
\setcitestyle{numbers,sort&compress}


\AtBeginDocument{%
  }

\copyrightyear{2026}
\acmYear{2026}
\setcopyright{cc}
\setcctype{by}
\acmConference[MM '26]{Proceedings of the 34th ACM International Conference on Multimedia}{November 10--14, 2026}{Rio de Janeiro, Brazil}
\acmBooktitle{Proceedings of the 34th ACM International Conference on Multimedia (MM '26), November 10--14, 2026, Rio de Janeiro, Brazil}
\acmDOI{10.1145/3767308.3836203}
\acmISBN{979-8-4007-2213-4/2026/11}
\begin{document}

\title{Agentic Tool-Augmented Reasoning for Explainable Image Forgery Detection}

\author{Zhiya Tan}
\orcid{0009-0004-2433-1389}
\authornote{This work was done when Zhiya Tan was an intern at Ant Digital Technologies, Ant Group. Project lead: Changtao Miao.}
\affiliation{%
  \department{CCDS}
  \institution{Nanyang Technological University}
  \city{Singapore}
  \country{Singapore}}
\affiliation{%
  \department{Ant Digital Technologies}
  \institution{Ant Group}
  \city{Singapore}
  \country{Singapore}}
\email{zhiya001@e.ntu.edu.sg}

\author{Jing Huang}
\orcid{0009-0005-5651-9033}
\affiliation{%
  \department{Ant Digital Technologies}
  \institution{Ant Group}
  \city{Singapore}
  \country{Singapore}}
\email{jh.jj@antgroup.com}

\author{Changtao Miao}
\orcid{0000-0002-7634-9992}
\affiliation{%
  \department{Ant Digital Technologies}
  \institution{Ant Group}
  \city{Hangzhou}
  \country{China}}
\email{miaoct1024@gmail.com}

\author{Lin Tan}
\orcid{0000-0003-4164-0672}
\affiliation{%
  \institution{Singapore University of Technology and Design}
  \city{Singapore}
  \country{Singapore}}
\email{lin_tan@sutd.edu.sg}

\author{Xin Zhang}
\orcid{0000-0002-6455-047X}
\affiliation{%
  \department{IAIC}
  \institution{Agency for Science, Technology and Research (A*STAR)}
  \city{Singapore}
  \country{Singapore}}
\email{zhangx7@a-star.edu.sg}

\author{Weiwei Feng}
\orcid{0000-0002-8761-0375}
\affiliation{%
  \department{Ant Digital Technologies}
  \institution{Ant Group}
  \city{Hangzhou}
  \country{China}}
\email{fengww@mail.ustc.edu.cn}

\author{Jianshu Li}
\orcid{0000-0001-8554-6886}
\affiliation{%
  \department{Ant Digital Technologies}
  \institution{Ant Group}
  \city{Singapore}
  \country{Singapore}}
\email{jianshu.l@antgroup.com}

\author{Joey Tianyi Zhou}
\orcid{0000-0002-4675-7055}
\authornote{Corresponding author.}
\affiliation{%
  \institution{Singapore Management University}
  \city{Singapore}
  \country{Singapore}}
\affiliation{%
  \department{IAIC}
  \institution{Agency for Science, Technology and Research (A*STAR)}
  \city{Singapore}
  \country{Singapore}}
\email{tyzhou@smu.edu.sg}

\renewcommand{\shortauthors}{Zhiya Tan et al.}

%%
%% Abstract
\begin{abstract}
\input{Content/Abs}
\end{abstract}

%%
%% Keywords.


\maketitle

%% Figure 1 -- Teaser
\begin{figure}[t]
  \centering
  \includegraphics[width=.92\columnwidth]{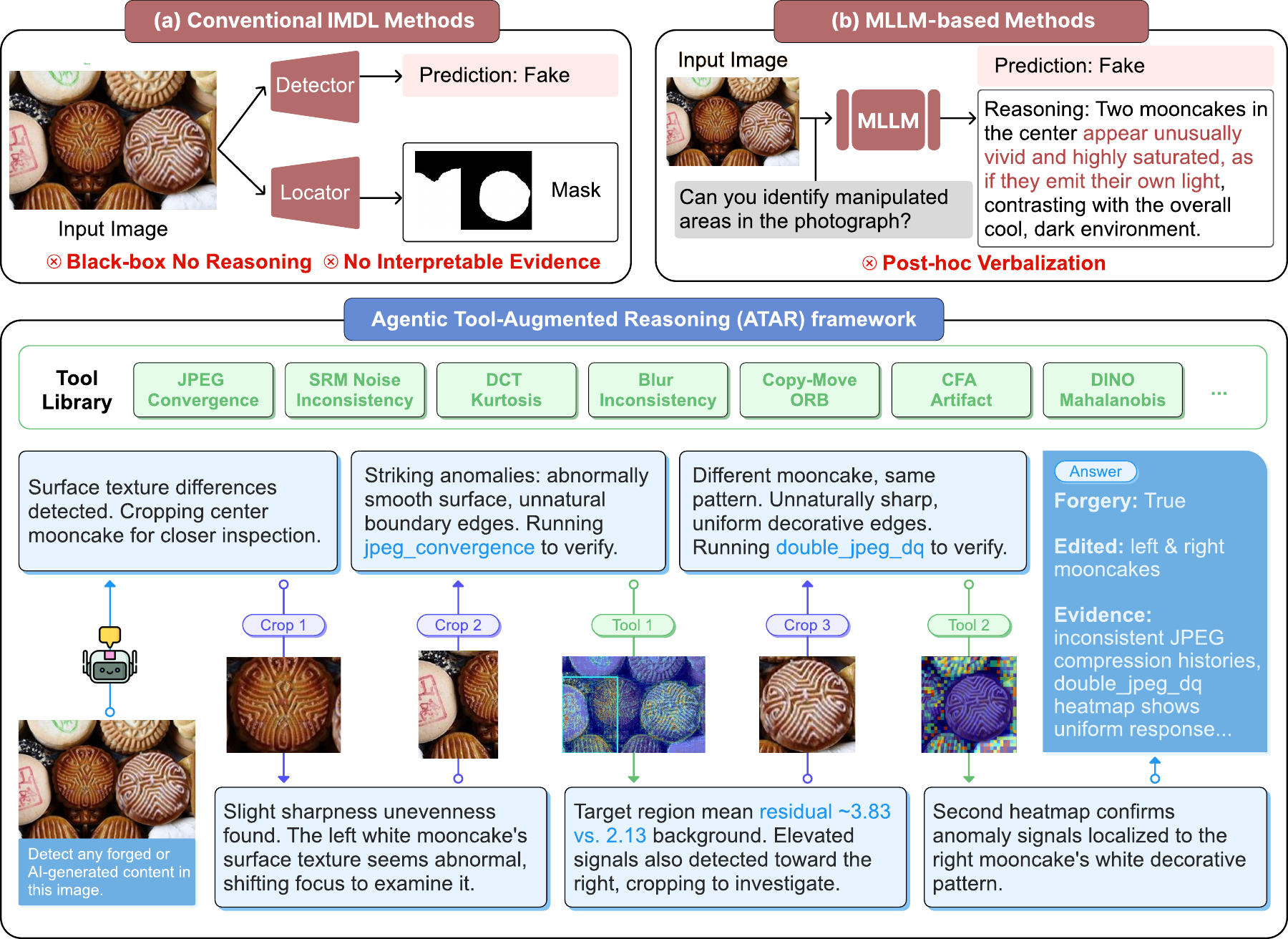}
  \caption{Three forgery detection paradigms. (a)~Conventional IMDL methods produce pixel masks but lack logical reasoning and interpretable evidence. (b)~MLLM-based methods generate post-hoc verbalizations of pre-determined detection results rather than genuine reasoning. (c)~ATAR autonomously invokes forensic tools and reasons from their outputs to produce grounded conclusions.}
  \Description{Comparison of three forgery detection paradigms.}
  \vspace{-\baselineskip}
  \label{fig:teaser}
\end{figure}

%% Figure 2 -- Method Main Figure (double-column)
\begin{figure*}[t]
  \centering
  \includegraphics[width=.94\textwidth]{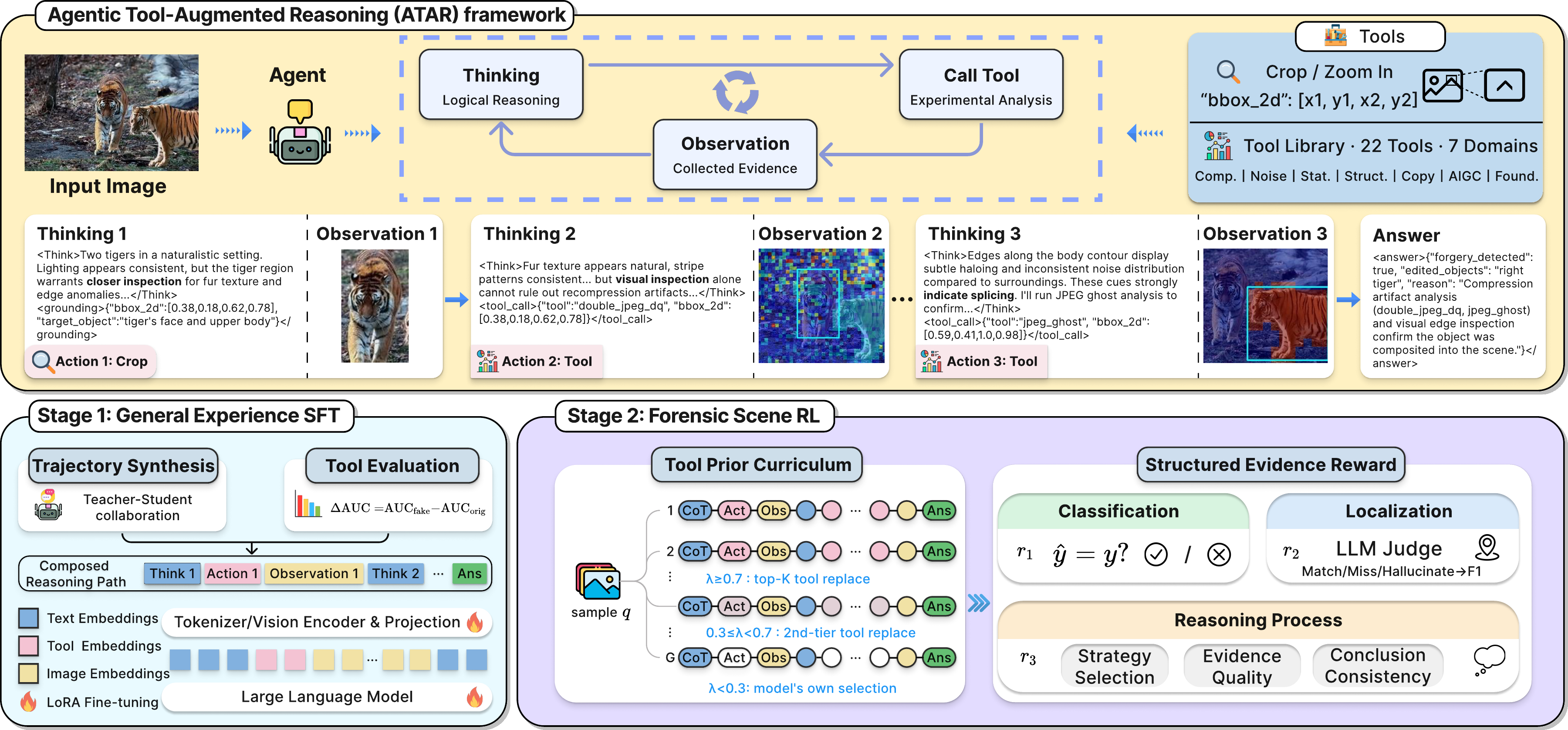}
  \caption{Overview of ATAR. \textbf{Top:} multi-turn reasoning chain. \textbf{Bottom:} two-stage training pipeline (Stage~1: General Experience SFT; Stage~2: Forensic Scene RL).}
  \Description{Overview of the ATAR method showing the reasoning chain and two-stage training pipeline.}
  \label{fig:method}
\end{figure*}

%% Introduction
\input{Content/intro_v2}

%% Related Work
\input{Content/related_work_en}

%% Method
\input{Content/method_v2.tex}

%% Experiments
\input{Content/experiment_main_en}

%% Conclusion
\input{Content/conclusion}

%%
%% Acknowledgments
\begin{acks}
This research is supported by the National Research Foundation, Singapore,
through its AI Singapore Programme (AISG Award No.~AISG3-RP-2024-033), and by
the Japan Science and Technology Agency and the Agency for Science, Technology
and Research under the Japan-Singapore Joint Call (Project No.~R24I6IR133).
This work also received support from Ant Digital Technologies, Ant Group during
Zhiya Tan's internship.
\end{acks}

%%
%% Bibliography
%% Manually start the final right-hand column at reference 62, as permitted by
%% the MM final-preparation instructions, to keep the last page balanced.
\newcounter{acmrefcounter}
\let\acmoriginalbibitem\bibitem
\renewcommand{\bibitem}{%
  \stepcounter{acmrefcounter}%
  \ifnum\value{acmrefcounter}=62\relax\vfill\eject\fi
  \acmoriginalbibitem}
\bibliographystyle{ACM-Reference-Format}
\bibliography{sample-base}

\end{document}

%% file: Content/Abs.tex
Conventional image forgery detection methods produce only binary scores or pixel-level masks without interpretable evidence, while recent multimodal large language model (MLLM)-based approaches generate textual explanations that are merely post-hoc verbalizations of pre-determined classification results rather than products of genuine reasoning.
Inspired by the forensic workflow of human judicial experts---``experimental analysis -- logical reasoning -- scientific evidence''---we propose an \textbf{A}gentic \textbf{T}ool-\textbf{A}ugmented \textbf{R}easoning (\textbf{ATAR}) framework for explainable image forgery detection, integrating 22 specialized forensic tools spanning seven complementary domains to autonomously detect, localize, and explain image forgeries through multi-turn reasoning.
Concretely, we propose a \textbf{Dual-Stream Forensic Reasoning} paradigm to emulate the experimental analysis process of forensic experts: (1) a high-level semantic anomaly path, which magnifies suspicious regions for fine-grained inspection; and (2) a low-level forgery artifact path, which invokes forensic tools to extract objective artifactual evidence.
To this end, we further propose a \textbf{Forensics Curriculum Learning} training strategy. First, during the General Experience SFT stage, an automated teacher--student mentoring pipeline is designed to synthesize multi-turn tool-usage reasoning trajectories. Subsequently, during the Forensic Scene RL stage, a Tool Prior Curriculum is introduced to guide early tool exploration and progressively transfer control to the agent, while a Structured Evidence Reward provides fine-grained process-level supervision.
Experiments across IMDL, Deepfake detection, DMDL, and AIGC detection show that ATAR achieves 78.5\% average image-level F1 on six zero-shot IMDL benchmarks, surpassing the strongest MLLM baseline by 11.8 percentage points, and remains competitive with specialized detectors on other tasks, while producing substantially more faithful and grounded explanations.

%% file: Content/intro_v2.tex
\section{Introduction}

The rapid advancement of generative AI~\cite{ddpm,ldm} and image editing techniques
has led to a steady increase in both the volume and realism of forged images,
posing growing threats to news credibility, forensic investigation,
and financial identity verification~\cite{forensicsoverview,deepfakesurvey}.
Effective forgery detection requires classifying
whether an image has been tampered with,
localizing the manipulated regions,
and providing interpretable reasoning for the decision.

% ---- P2: Existing methods ----
Despite significant progress,
existing methods each address only part of these requirements,
as shown in Figure~\ref{fig:teaser}.
Conventional image manipulation detection and localization (IMDL)
methods~\cite{mvssnet,catnet,trufor}
capture low-level forgery cues
such as boundary artifacts,
JPEG compression inconsistencies,
and frequency-domain anomalies~\cite{noiseprint,freqaware},
yet their black-box nature provides no interpretable reasoning
and cannot produce traceable evidence for the decision.
Conversely, multimodal large language model (MLLM)-based
approaches~\cite{forgerygpt,fakeshield,ffaa}
combine detection modules with language generation~\cite{llava,qwen2vl}
to produce both predictions and textual explanations.
However, detection decisions are produced by the modules independently,
and the language model merely verbalizes
these pre-determined outcomes
rather than reasoning from the underlying evidence.
Moreover, the visual encoders
lack perception of low-level traces~\cite{clip},
forcing the model to produce plausible yet ungrounded explanations
when no obvious semantic flaw exists.
In both cases, the core limitation is the same:
analysis and reasoning are decoupled,
preventing the system from producing grounded explanations.

% ---- P3: Judicial forensic workflow ----
In fact, a well-established methodology already exists in judicial forensic practice,
where human experts follow a structured three-stage workflow~\cite{faridphoto,forensicworkflow}:
\textbf{experimental analysis}, \textbf{logical reasoning}, and \textbf{scientific evidence}.
During experimental analysis,
the expert first inspects the image
for semantic anomalies~\cite{physicforensic},
then magnifies suspicious regions for closer examination.
When no semantic flaw is apparent,
specialized instruments are applied~\cite{ela,srm,crfanalysis}
to reveal signal-level traces invisible to the naked eye.
The expert then performs logical reasoning
over the collected experimental results to form a judgment,
and presents the final conclusion as scientific evidence
with a traceable chain of support.
This close integration of analysis, reasoning, and evidence
is absent from existing automated methods.

% ---- P4: ATAR framework ----
Motivated by this forensic workflow,
we propose \textbf{A}gentic \textbf{T}ool-\textbf{A}ugmented \textbf{R}easoning (\textbf{ATAR})
to reproduce the complete
``experimental analysis -- logical reasoning -- scientific evidence''
workflow through multi-turn interactive reasoning.
For experimental analysis,
we note that the expert's examination involves
two complementary types of forgery cues,
namely high-level semantics and low-level artifacts,
and accordingly design a \textbf{Dual-Stream Forensic Reasoning} paradigm:
the high-level semantic anomaly path
magnifies suspicious regions for fine-grained inspection,
while the low-level forgery artifact path
invokes 22 unsupervised tools
spanning seven complementary domains
to extract objective evidence.
For logical reasoning,
the MLLM interprets observations from both paths
across multiple turns,
autonomously deciding whether to initiate further analysis
based on intermediate results.
The final output constitutes scientific evidence:
a classification decision, localization of forged regions,
and a traceable chain of evidence supporting the verdict.

% ---- P5: FCL training ----
To train the model to master this workflow,
we propose \textbf{Forensics Curriculum Learning}.
In the \textbf{General Experience SFT} stage,
an automated teacher--student pipeline
performs \textbf{Forensic Reasoning Trajectory Synthesis},
converting raw forgery datasets
into multi-turn reasoning trajectories that capture genuine exploratory reasoning
with each step grounded in tool outputs or semantic observations.
In the subsequent Forensic Scene reinforcement learning (RL) stage,
since the post-SFT model cannot yet reliably select
the right tool from 22 candidates,
a \textbf{Tool Prior Curriculum}
provides effective tool selections early in training
and progressively removes this guidance,
enabling the model to transition
from assisted to independent tool use.
Meanwhile, to ensure that the reasoning process itself
yields reliable scientific evidence,
a \textbf{Structured Evidence Reward}
decomposes supervision
along classification, localization, and reasoning quality
into independently verifiable items,
ensuring that each dimension of the reasoning process
receives targeted gradient signals.

Extensive experiments on IMDL~\cite{casia,columbia},
Deepfake detection~\cite{genimage},
document manipulation detection and localization (DMDL)~\cite{docforgebench},
and AI-generated content (AIGC) detection demonstrate that
ATAR achieves state-of-the-art results:
78.5\% image-level F1 and 56.4\% pixel-level F1 on six zero-shot IMDL benchmarks
(+5.6 percentage points (pp) and +6.0\,pp over TruFor),
99.9\% F1 on Deepfake detection,
and 87.0\% F1 on AIGC detection without any AIGC training data,
while reducing reasoning hallucination to 7.2\% on true positives.
Our main contributions are as follows.

\begin{itemize}
\item We propose \textbf{ATAR} with a \textbf{Dual-Stream Forensic Reasoning} paradigm
  that reproduces the judicial forensic workflow
  through multi-turn interactive reasoning,
  enabling MLLMs to autonomously invoke forensic tools
  and produce scientific evidence.

\item We design \textbf{Forensic Reasoning Trajectory Synthesis},
  an automated teacher--student pipeline
  that converts raw forgery datasets into multi-turn tool-augmented reasoning trajectories
  for the \textbf{General Experience SFT} stage.

\item We propose \textbf{Forensic Scene RL}
  with a \textbf{Tool Prior Curriculum} that guides early tool exploration
  and progressively transfers control to the agent,
  and a \textbf{Structured Evidence Reward} that decomposes supervision
  into independently verifiable items for classification, localization, and reasoning quality.
\end{itemize}

%% file: Content/related_work_en.tex
\section{Related Work}

\subsection{Image Forgery Detection and MLLMs}
\label{sec:rw_forgery}

Deep learning methods detect forgeries by capturing low-level traces:
boundary and noise-view artifacts~\cite{mantranet,mvssnet,noiseprint},
JPEG compression inconsistencies~\cite{catnet},
high-frequency object-level features~\cite{objectformer},
and learned noise fingerprints~\cite{trufor}.
For AI-generated images,
methods exploit shared generator artifacts~\cite{cnndetection},
diffusion reconstruction error~\cite{dire},
or frozen CLIP-ViT features~\cite{univfd}.
These methods achieve high accuracy but provide no interpretable evidence.

To address interpretability,
recent works apply multimodal large language models (MLLMs)~\cite{llava,qwen2vl} to forgery detection.
Several approaches augment MLLMs with dedicated forensic modules---mask-aware extractors~\cite{forgerygpt},
domain-guided detectors~\cite{fakeshield,x2dfd},
or trace encoders~\cite{forgerysleuth}---to
fuse semantic and low-level cues.
Others reformulate detection as a reasoning task
via dual-branch encoders~\cite{fakereasoning},
hypothetical prompting~\cite{ffaa},
or GRPO-based reasoning training~\cite{thinkfake}.
However, MLLM visual encoders such as CLIP~\cite{clip} are designed for high-level semantics
and cannot perceive low-level traces like JPEG artifact misalignment or noise anomalies.
Without genuine low-level evidence,
the model may hallucinate unfounded explanations.
No existing method bridges this gap with external forensic tools.

\subsection{Tool-Augmented Reasoning}
\label{sec:rw_tool}

A growing line of work enables MLLMs to actively interact with images during multi-turn reasoning.
Several methods train models to crop, zoom, and re-examine regions
via RL-incentivized interleaved reasoning~\cite{deepeyes,deepeyesv2,minio3,simpleo3},
progressive visual grounding curricula~\cite{vthinker},
RL-driven external tool invocation~\cite{openthinkimg},
or plug-and-play visual search~\cite{insighto3}.
For tool-augmented LLMs more broadly,
ReAct~\cite{react} interleaves reasoning and action to reduce hallucination,
and T$^2$Agent~\cite{t2agent} coordinates modular tools with MCTS
for multimodal misinformation detection.

For training,
GRPO~\cite{deepseekmath} replaces the critic with within-group normalization,
and curriculum strategies~\cite{scafgrpo,ruscarl} provide tiered guidance
that is progressively removed during exploration.
Applying these to forgery detection poses two challenges:
existing rewards use a single score
that conflates classification, localization, and evidence quality;
and with a large tool space, random exploration is inefficient,
while current curricula operate at the prompt level rather than tool selection.

%% file: Content/method_v2.tex
\section{Method}

ATAR reproduces the judicial forensic workflow
of ``experimental analysis -- logical reasoning -- scientific evidence''
by equipping an MLLM with multi-turn reasoning.
For \textbf{experimental analysis},
the model autonomously decides, at each turn,
whether to magnify a region for semantic inspection
or to invoke a forensic tool for signal-level evidence.
For \textbf{logical reasoning},
it interprets observations across turns,
deciding whether to initiate further analysis
or to form a judgment.
The output constitutes \textbf{scientific evidence}:
a classification verdict, localization of forged regions,
and a traceable chain of evidence supporting the conclusion.
Section~\ref{sec:framework} details the \textbf{Dual-Stream Forensic Reasoning} paradigm that instantiates this workflow;
Sections~\ref{sec:coldstart}--\ref{sec:reward} present \textbf{Forensics Curriculum Learning}, comprising trajectory synthesis, tool prior curriculum, and structured evidence reward.

\subsection{Dual-Stream Forensic Reasoning}
\label{sec:framework}

Given an input image $I$, the model produces
an authenticity verdict $\hat{y} \in \{\text{authentic}, \text{forged}\}$,
textual descriptions of edited objects $\{\hat{e}_1, \dots, \hat{e}_P\}$
(e.g., ``the tiger in the center was spliced''),
and an interpretable reasoning trajectory $o$
that constitutes the scientific evidence.
During reasoning, the model can crop a specified region
for magnified local details,
or invoke analysis tools from a library
$\mathcal{T} = \{T_1, \dots, T_{22}\}$
that produce heatmaps at the same spatial resolution as $I$.

We categorize forgery cues into two top-level types~\cite{skyra},
each determining a distinct reasoning path---together
forming the \textbf{Dual-Stream Forensic Reasoning} paradigm
that emulates the experimental analysis process of forensic experts.
\textbf{High-level Semantics}
refers to forgeries that violate physical laws or commonsense logic.
The model identifies these by cropping and magnifying
the suspicious region for semantic analysis,
paralleling the analyst's visual inspection stage.
\textbf{Low-level Artifacts}
are imperceptible anomalies in signal-level patterns
that appear semantically flawless.
The model must invoke analysis tools for verification,
paralleling the analyst's instrumental analysis stage.
The two paths are not mutually exclusive;
the model may switch paths
when one reasoning direction yields insufficient evidence.

The tool library provides 22 unsupervised tools
spanning seven complementary domains:
compression artifacts, sensor noise,
statistical and frequency distributions, structural inconsistencies,
copy-move duplications, AI-generated-specific patterns,
and foundation-model-based anomaly detection
(DINOv3~\cite{dinov3} Layer~4 features
with principal component analysis (PCA)-reduced Mahalanobis distance).
Each reasoning turn contains text and one action tag~\cite{react}:
\texttt{<grounding>} specifies a region
via a normalized bounding box $[x_1, y_1, x_2, y_2] \in [0,1]^4$
and a target object description, triggering a crop;
\texttt{<tool\_call>} invokes a tool;
and \texttt{<answer>} delivers the final verdict
along with the traceable chain of evidence.
The environment returns results as input for the next turn.
This multi-turn loop instantiates the three-stage
judicial forensic workflow
as illustrated in Figure~\ref{fig:method} (Forensic Reasoning Loop):
\texttt{<grounding>} and \texttt{<tool\_call>}
perform experimental analysis,
the reasoning between turns performs logical reasoning,
and the final \texttt{<answer>} with its accumulated trajectory
constitutes the scientific evidence.

ATAR is trained with a two-stage pipeline
that we term Forensics Curriculum Learning
as shown in Figure~\ref{fig:method}:
the General Experience SFT stage synthesizes reasoning trajectories
via a teacher--student mechanism (Section~\ref{sec:coldstart}),
while the Forensic Scene RL stage introduces
a Tool Prior Curriculum (Section~\ref{sec:curriculum})
and a Structured Evidence Reward (Section~\ref{sec:reward}).
The figure also illustrates an example
where the model crops a suspicious tiger region,
invokes \texttt{double\_jpeg\_dq} and \texttt{jpeg\_ghost}~\cite{dso1},
and concludes from converging evidence
that the tiger was spliced.

\begin{figure}[t]
  \centering
  \includegraphics[width=.92\columnwidth]{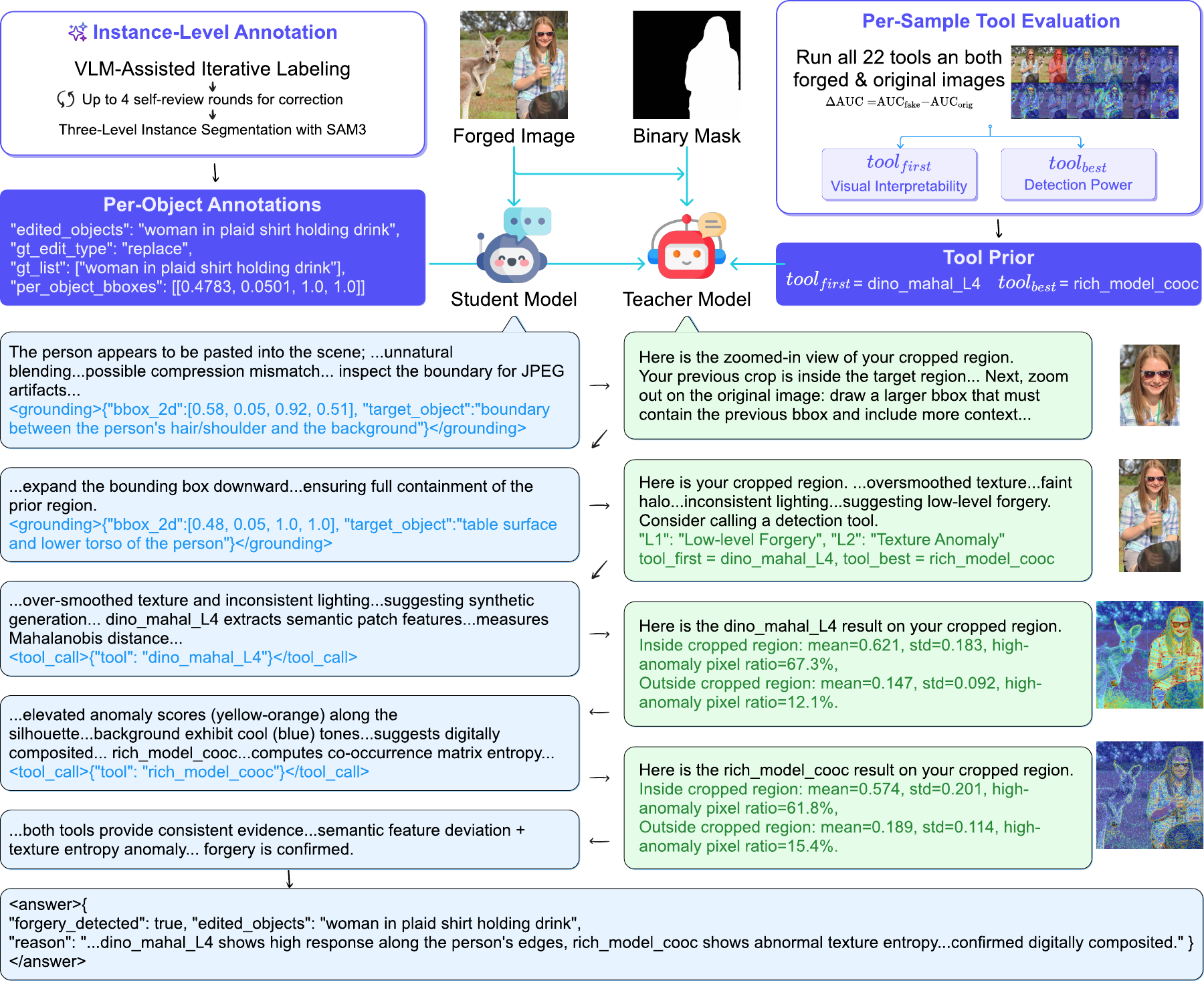}
  \caption{General Experience trajectory synthesis pipeline.
  Left: instance-level annotation.
  Right: per-sample tool evaluation.
  Bottom: teacher-guided trajectory generation,
  where the student explores without ground truth and the teacher provides corrective guidance.}
  \Description{General Experience SFT data synthesis pipeline with annotation, tool evaluation, and trajectory generation.}
  \label{fig:framework}
\end{figure}

\subsection{Forensic Reasoning Trajectory Synthesis}
\label{sec:coldstart}

Existing forgery datasets lack multi-turn reasoning annotations,
and the cost of manually annotating
multi-turn tool-interleaved trajectories is prohibitive.
As shown in Figure~\ref{fig:framework},
we design an automated pipeline that converts raw forgery datasets
into tool-augmented reasoning trajectories
through four stages.

\subsubsection{\textbf{Instance Annotation}}
\label{sec:instance}

Raw datasets provide only binary masks.
We use an MLLM-assisted iterative scheme to decompose them
into per-object instance masks with semantic labels,
bounding boxes, and tampering types,
with Segment Anything Model (SAM)~\cite{sam3} invoked for instance separation when needed.

\subsubsection{\textbf{Per-Sample Tool Evaluation}}
\label{sec:tooleval}

Since no single tool is universally optimal,
we run all 22 tools on each forged image
and its pristine original (available for all training sets)
and compute pixel-level area under the curve (AUC) against the ground-truth mask.
The difference
$\Delta\text{AUC} = \text{AUC}_{\text{fake}} - \text{AUC}_{\text{orig}}$
measures net detection effectiveness,
with the original-image baseline eliminating the tool's inherent bias.
A tool with high $\Delta\text{AUC}$ may still produce
heatmaps where the forged region is not visually distinguishable.
We therefore select two complementary tools per sample.
$\text{tool\_first}$ selects the tool whose heatmap
is easiest for the MLLM to interpret
by maximizing Cohen's $d$~\cite{cohen2013statistical} between tampered and background intensities,
clamped to non-negative and scaled by $\Delta\text{AUC}$
so that only tools with sufficient detection quality are considered.
$\text{tool\_best}$ selects the most accurate tool
by maximizing $\Delta\text{AUC}$ directly,
regardless of visual clarity.
A Top-$K$ ($K{=}3$) optimal tool set is also generated per sample
for curriculum guidance and reward computation.
Samples whose highest $\Delta\text{AUC}$ falls below 0.15
are excluded from tool-augmented training.

\input{Content/table_tool_complementarity}

Table~\ref{tab:tool_complementarity} reveals
that optimal tool domains shift systematically
with the underlying forgery technique:
JPEG-based splicing (AutoSplice) concentrates 75.7\% on compression tools,
whereas adversarial composites (FantasticReality) primarily trigger noise tools (29.7\%).
Every tool domain leads on at least one dataset,
confirming the library is both necessary and non-redundant.

\subsubsection{\textbf{Teacher-Guided Trajectory Generation}}
\label{sec:trajectory}

Both the student and teacher are Qwen3-VL-235B-A22B~\cite{qwen3vl}.
The student explores images without ground truth;
the teacher holds annotations and guides exploration
through three mechanisms, as shown in Figure~\ref{fig:framework}.
After each student crop,
the teacher matches it against uncovered ground-truth masks
by mask coverage (fraction of the ground-truth mask covered by the crop)
and overlap ratio (fraction of the crop occupied by the mask).
The result is classified as
hit (coverage ${\geq}0.5$, overlap ${\geq}0.12$),
relevant (coverage ${\geq}0.1$, overlap ${\geq}0.05$),
or irrelevant,
with corresponding feedback
(confirmation, refinement guidance, or redirection).
The teacher tracks covered instances
to ensure all tampered regions are discovered,
and escalates prompt intensity across turns,
from directional hints to explicit target specification.
For low-level cues, the teacher recommends
$\text{tool\_first}$ and $\text{tool\_best}$ sequentially;
when a crop yields no anomaly,
the student explicitly reflects on its failed hypothesis,
producing trajectories with complete
hypothesis, verification, rejection, and adjustment cycles.
When the student repeatedly fails,
the teacher injects ground-truth locations as a fallback.
All trajectories pass automated verification
that checks full instance coverage, action-tag well-formedness,
and tool-call validity before inclusion in the SFT set.

\subsubsection{\textbf{General Experience SFT}}
\label{sec:sft}

We apply Low-Rank Adaptation (LoRA)~\cite{lora} to all linear layers of Qwen3-VL-8B~\cite{qwen3vl}
with the vision encoder frozen and multi-modal projector unfrozen
to adapt visual-textual alignment to domain-specific inputs
such as heatmaps and cropped forensic views.
The loss is computed only on model-generated tokens,
excluding environment-returned observations.

In the Forensic Scene RL stage,
the SFT model has acquired the multi-turn reasoning format
but cannot yet autonomously select optimal tools.
We employ GRPO~\cite{deepseekmath} for policy optimization,
facing two challenges:
sound logical reasoning presupposes effective experimental analysis,
so we must first guide efficient tool exploration
over a large 22-tool action space (Section~\ref{sec:curriculum});
with tool selection addressed,
effective supervision of the reasoning process itself is needed
to produce reliable scientific evidence (Section~\ref{sec:reward}).

% ====================================================================
% Table: Merged IMDL -- placed here so table* appears on Experiments page
% ====================================================================
\begin{table*}[t]
\centering
\vspace{-2mm}
\caption{IMDL benchmark results on six zero-shot datasets. iF1/ACC measure image-level classification; pF1/IoU measure pixel-level localization. Methods above the first rule are specialized detectors; below are MLLM-based and general-purpose MLLMs. 
\vspace{-2mm}
\textbf{Bold} = best, \underline{underline} = second.}
\label{tab:imdl}
\renewcommand{\arraystretch}{0.95}
\setlength{\tabcolsep}{1pt}
\resizebox{\textwidth}{!}{%
\scriptsize
\begin{tabular}{l cccc cccc cccc cccc cccc cccc cccc}
\toprule
 & \multicolumn{4}{c}{CASIA v1+}
 & \multicolumn{4}{c}{CocoGlide}
 & \multicolumn{4}{c}{Coverage}
 & \multicolumn{4}{c}{Korus}
 & \multicolumn{4}{c}{Columbia}
 & \multicolumn{4}{c}{NIST16}
 & \multicolumn{4}{c}{\textbf{Avg}} \\
\cmidrule(lr){2-5} \cmidrule(lr){6-9} \cmidrule(lr){10-13} \cmidrule(lr){14-17} \cmidrule(lr){18-21} \cmidrule(lr){22-25} \cmidrule(lr){26-29}
\multicolumn{1}{c}{Method}
 & iF1 & ACC & pF1 & IoU
 & iF1 & ACC & pF1 & IoU
 & iF1 & ACC & pF1 & IoU
 & iF1 & ACC & pF1 & IoU
 & iF1 & ACC & pF1 & IoU
 & iF1 & ACC & pF1 & IoU
 & iF1 & ACC & pF1 & IoU \\
\midrule
% --- Specialized detection methods (optimal F1 threshold) ---
MVSS-Net~\cite{mvssnet} (\textit{ICCV'21})
 & .734 & .757 & .471 & .397
 & .665 & .560 & .492 & .372
 & .669 & .550 & .467 & .373
 & .632 & .539 & .181 & .120
 & .749 & .672 & .670 & .576
 & \underline{.682} & .538 & .299 & .216
 & .689 & .603 & .430 & .342 \\
TruFor~\cite{trufor} (\textit{CVPR'23})
 & .732 & .652 & .544 & .464
 & .685 & .576 & \underline{.516} & .405
 & \underline{.681} & .570 & .450 & .346
 & \underline{.679} & .559 & .292 & .217
 & \textbf{.975} & \textbf{.975} & \textbf{.833} & \textbf{.769}
 & .620 & .466 & .391 & .297
 & .729 & .633 & \underline{.504} & \underline{.416} \\
ForMa~\cite{forma} (\textit{SPL'25})
 & .781 & .769 & \underline{.629} & \underline{.574}
 & .668 & .502 & .453 & .362
 & .669 & .525 & .487 & .409
 & .670 & .509 & \underline{.304} & \underline{.235}
 & .683 & .650 & .001 & .000
 & .570 & .487 & .055 & .030
 & .674 & .574 & .322 & .268 \\
SparseViT~\cite{sparsevit} (\textit{AAAI'25})
 & .697 & .535 & .153 & .088
 & .667 & .500 & .355 & .253
 & .667 & .500 & .197 & .112
 & .667 & .500 & .103 & .057
 & .607 & .435 & .000 & .000
 & .563 & .392 & \textbf{.552} & \textbf{.519}
 & .645 & .477 & .227 & .172 \\
\midrule
% --- MLLM-based detection methods ---
FakeShield~\cite{fakeshield} (\textit{ICLR'25})
 & \underline{.891} & \underline{.883} & .594 & .538
 & .669 & .506 & \textbf{.521} & \textbf{.428}
 & .619 & .525 & .242 & .212
 & .467 & .559 & .125 & .102
 & .920 & .920 & .744 & .658
 & .438 & \underline{.554} & .260 & .228
 & .667 & .658 & .414 & .361 \\
SIDA-7B~\cite{sida} (\textit{CVPR'25})
 & .437 & .595 & .033 & .019
 & .483 & .639 & .081 & .054
 & .000 & .500 & .125 & .072
 & .018 & .505 & .065 & .037
 & .301 & .590 & .471 & .368
 & .060 & .454 & .149 & .092
 & .217 & .547 & .154 & .107 \\
\midrule
% --- General-purpose MLLMs (zero-shot) ---
GPT-5.4
 & .571 & .700 & .263 & .178
 & .369 & .590 & .449 & .378
 & .295 & .570 & .288 & .189
 & .140 & .510 & .256 & .194
 & .828 & .850 & .616 & .517
 & .233 & .540 & .115 & .091
 & .406 & .627 & .331 & .258 \\
Gemini-2.5-Pro
 & .698 & .741 & .294 & .216
 & .540 & .664 & .449 & .381
 & .314 & .520 & .131 & .086
 & .396 & \underline{.598} & .143 & .101
 & .866 & .876 & .453 & .363
 & .279 & .510 & .455 & .347
 & .516 & .652 & .321 & .249 \\
Qwen3-VL-8B
 & .333 & .563 & .199 & .138
 & .211 & .540 & .489 & .386
 & .076 & .510 & .296 & .215
 & .127 & .530 & .246 & .170
 & .869 & .870 & .631 & .540
 & .121 & .471 & .211 & .194
 & .290 & .581 & .345 & .274 \\
\midrule
% --- Ours ---
\rowcolor{gray!10} ATAR (sft only)
 & .882 & .875 & .520 & .468
 & \underline{.694} & \underline{.686} & .421 & .340
 & .673 & \underline{.641} & \underline{.507} & \underline{.452}
 & .643 & .591 & .260 & .187
 & .951 & .950 & .742 & .679
 & .616 & .520 & .359 & .288
 & \underline{.743} & \underline{.711} & .468 & .402 \\
\rowcolor{gray!10} \textbf{ATAR (Ours)}
 & \textbf{.929} & \textbf{.924} & \textbf{.681} & \textbf{.624}
 & \textbf{.718} & \textbf{.711} & .503 & \underline{.422}
 & \textbf{.715} & \textbf{.680} & \textbf{.545} & \textbf{.509}
 & \textbf{.696} & \textbf{.639} & \textbf{.348} & \textbf{.252}
 & \underline{.962} & \underline{.962} & \underline{.825} & \underline{.760}
 & \textbf{.690} & \textbf{.561} & \underline{.484} & \underline{.408}
 & \textbf{.785} & \textbf{.746} & \textbf{.564} & \textbf{.496} \\
\bottomrule
\end{tabular}%
}
\end{table*}

\subsection{Tool Prior Curriculum Learning}
\label{sec:curriculum}

With 22 candidates, the post-SFT model
frequently selects suboptimal tools.
An ill-chosen tool not only wastes a reasoning turn
but actively misleads the model,
which treats irrelevant highlights as evidence.
When most selections produce uninformative evidence,
the RL signal is too sparse for effective learning.
We address this through a Tool Prior Curriculum~\cite{ruscarl}
that injects optimal tool priors early and progressively removes them.

When the model selects a tool on the low-level artifacts path,
the environment replaces it without the model's knowledge.
For each input, GRPO samples $G$ trajectories,
each assigned a curriculum intensity $\lambda(i, t)$
that determines the replacement tier:
$\lambda \geq 0.7$ triggers replacement with a Top-$K$ optimal tool,
$0.3 \leq \lambda < 0.7$ with a second-tier tool
(ranked 4th--7th by $\Delta\text{AUC}$),
and $\lambda < 0.3$ retains the model's own selection.

This mixing creates within-group reward variance
essential for meaningful GRPO gradients.
High-level-semantics-path trajectories bypass the curriculum entirely.
The intensity is defined as
\begin{equation}
\lambda(i, t) = \underbrace{\frac{G - i}{G - 1}}_{\text{within-group diversity}}
\cdot \underbrace{\frac{1}{1 + \exp(\alpha(t - t_0))}}_{\text{global decay over training}}
\label{eq:curriculum}
\end{equation}

We use GRPO with within-group normalization
and asymmetric clipping~\cite{scafgrpo} ($\epsilon^{-} < \epsilon^{+}$),
granting larger updates to positive-advantage trajectories.
\begin{align}
\mathcal{L}_{\text{GRPO}} = &-\mathbb{E}_{q} \!\left[ \frac{1}{G}
  \sum_{i=1}^{G} \min\!\Big( \rho_i \hat{A}_i,\,
  \text{clip}(\rho_i, 1\!-\!\epsilon^{-}\!, 1\!+\!\epsilon^{+})
  \hat{A}_i \Big) \right] \nonumber \\
  &+ \beta \, D_{\text{KL}}(\pi_\theta \| \pi_{\text{ref}})
\label{eq:grpo}
\end{align}
where $\rho_i = \pi_\theta(o_i|q) / \pi_{\text{old}}(o_i|q)$
and $\hat{A}_i = (r_i - \mu_G) / \sigma_G$.
When the curriculum triggers a replacement,
the tool name in the model's output is replaced
and the prompt sequence is re-tokenized
so subsequent turns observe the replaced tool as context.
High rewards from replaced tools produce gradient signals
that guide the model toward selecting those tools independently,
completing the transition as the curriculum decays
(ToolAcc rises from 41.8\% to 72.0\%).

\subsection{Structured Evidence Reward}
\label{sec:reward}

A model can arrive at the correct verdict
through flawed reasoning,
and a single scalar reward would reinforce such behavior
by conflating the final answer with the reasoning process.
To produce reliable scientific evidence,
we decompose the reward
into four independent components~\cite{rubrichub}
so that each ability is supervised separately.
Each is either computed by deterministic rules
or assessed by an LLM judge constrained to binary decisions.
\begin{equation}
r = w_1 \cdot r_1 + w_2 \cdot r_2 + w_3 \cdot r_3 + w_4 \cdot r_{\text{fmt}}
\label{eq:reward}
\end{equation}
$r_1 = \mathbf{1}[\hat{y} = y]$ rewards correct classification.
$r_{\text{fmt}}$ penalizes malformed action tags or invalid tool calls.
The remaining two components require more careful design.

Since the model outputs textual region descriptions $\hat{e}_i$
rather than pixel masks,
standard IoU cannot evaluate localization directly.
For $r_2$, we overlay the ground-truth mask on the image
and submit it together with the model's textual predictions
to the LLM judge,
which classifies each predicted region
as matched, missed, or hallucinated.
\begin{equation}
r_2 = \begin{cases}
\dfrac{2 \cdot \text{Prec} \cdot \text{Rec}}{\text{Prec} + \text{Rec}}
  & \text{forged} \\[4pt]
\mathbf{1}[P = 0] & \text{authentic}
\end{cases}
\label{eq:r2}
\end{equation}
Here, $P$ counts predicted regions, while $|\mathcal{M}|$ counts those
matched by the judge. Let $N_{\text{miss}}$ denote the ground-truth instances
that are not matched by any prediction. We compute
$\text{Prec}=|\mathcal{M}|/P$ and
$\text{Rec}=|\mathcal{M}|/(|\mathcal{M}|+N_{\text{miss}})$.

To further supervise the reasoning process itself,
$r_3 = (c_1 + c_2 + c_3) / 3$ evaluates
reasoning quality through three criteria.
$c_1$ and $c_2$ assess per-turn quality
for each matched object
under three scenarios corresponding to the reasoning path;
$c_3$ assesses the overall trajectory quality.

\textbf{Strategy Selection $c_1$.}
For low-level artifacts,
a deterministic rule checks whether the invoked tool
belongs to the Top-$K$ optimal set.
For high-level semantics,
the LLM judge verifies that the model identified
a genuine physical or commonsense violation
rather than fabricating one.
For authentic images,
it checks that at least one tool was invoked,
preventing the model from declaring authenticity
based solely on the absence of semantic anomalies.

\textbf{Evidence Quality $c_2$.}
For low-level artifacts,
the LLM judge checks whether heatmap descriptions
match actual tool outputs.
For high-level semantics,
it checks whether semantic observations are grounded in the image.
For authentic images,
it checks whether heatmaps were correctly interpreted as normal.

\textbf{Conclusion Consistency $c_3$.}
The LLM judge verifies that the final verdict
is consistent with accumulated evidence
and that contradictions are explicitly acknowledged
rather than ignored.

%% file: Content/table_tool_complementarity.tex
\begin{table}[t]
\centering
\caption{Domain-level optimal tool selection proportion (\%) across training datasets. Per-tool breakdown is provided in the appendix due to space constraints.}
\label{tab:tool_complementarity}
\renewcommand{\arraystretch}{0.94}
\resizebox{\columnwidth}{!}{%
\small
\setlength{\tabcolsep}{4pt}
\newcommand{\cmark}[2]{\cellcolor{blue!#1!white}{#2}}
\begin{tabular}{@{}l*{7}{c}@{}}
\toprule
 & \multicolumn{5}{c}{IMDL} & {Deepfake} & {DMDL} \\
\cmidrule(lr){2-6} \cmidrule(lr){7-7} \cmidrule(lr){8-8}
 & {AutoSplice} & {CASIA2} & {Fant.Reality} & {IMD2020} & {NeXT-rpl.} & {OpenForen.} & {DocTamper} \\
\midrule
Compress. (4)  & \cmark{62}{75.7} & \cmark{28}{33.6} & \cmark{6}{6.7}   & \cmark{12}{15.0} & \cmark{2}{2.4}   & \cmark{33}{40.7} & \cmark{1}{1.2} \\
Noise (4)      & \cmark{6}{7.7}   & \cmark{15}{18.6} & \cmark{24}{29.7} & \cmark{18}{22.5} & \cmark{25}{30.6} & \cmark{6}{7.7}   & \cmark{33}{39.8} \\
Freq. (3)      & \cmark{1}{0.8}   & \cmark{3}{3.9}   & \cmark{14}{16.9} & \cmark{6}{7.4}   & \cmark{6}{7.9}   & \cmark{2}{2.9}   & \cmark{3}{3.9} \\
Struct. (3)    & \cmark{1}{1.1}   & \cmark{6}{6.8}   & \cmark{8}{9.4}   & \cmark{6}{7.3}   & \cmark{11}{13.9} & \cmark{5}{6.0}   & \cmark{3}{4.2} \\
Copy-Move (1)  & \cmark{1}{0.9}   & \cmark{7}{8.8}   & \cmark{4}{5.2}   & \cmark{5}{5.7}   & \cmark{6}{7.6}   & \cmark{7}{9.2}   & \cmark{14}{16.8} \\
AI-Gen. (6)       & \cmark{11}{13.7} & \cmark{13}{16.5} & \cmark{13}{16.0} & \cmark{16}{20.1} & \cmark{25}{30.6} & \cmark{25}{31.1} & \cmark{19}{23.2} \\
Found.Mod. (1) & \cmark{1}{0.2}   & \cmark{9}{11.7}  & \cmark{13}{16.0} & \cmark{18}{22.0} & \cmark{6}{7.0}   & \cmark{2}{2.4}   & \cmark{9}{10.9} \\
\bottomrule
\end{tabular}%
}
\end{table}

%% file: Content/experiment_main_en.tex
\section{Experiments}
\label{sec:experiment}

\subsection{Experimental Setup}
\label{sec:exp_setup}

We train on three domains: image manipulation detection and localization (IMDL), Deepfake detection, and document manipulation detection and localization (DMDL), and additionally evaluate zero-shot AIGC detection.
SFT uses ${\sim}$41K multi-turn trajectories from seven datasets: CASIA2~\cite{casia}, IMD2020~\cite{imd2020}, FantasticReality~\cite{fantasticreality}, NeXT-IMDL~\cite{nextimdl}, AutoSplice~\cite{autosplice} (IMDL), OpenForensics~\cite{openforensics} (Deepfake), and DocTamper~\cite{doctamper} (DMDL).
RL uses NeXT-IMDL, OpenForensics, and DocTamper ({62K} samples).
In-domain testing follows official splits of these three datasets.
Out-of-domain IMDL sets with pixel masks (CASIAv1+~\cite{casia}, Columbia~\cite{columbia}, NIST16~\cite{nist16}, Coverage~\cite{coverage}, Korus~\cite{korus}, CocoGlide~\cite{trufor}) evaluate classification and localization; document forgery (T-SROIE~\cite{tornes2023receipt}, FSTS-1.5k~\cite{yu2025toward}) evaluates localization only; GenImage++~\cite{genimage} (5 generator families) evaluates AIGC classification only.
Classification uses image-level Accuracy and F1. For localization, ATAR's textual region descriptions are converted to pixel masks via Grounded-SAM~\cite{groundedsam}, then evaluated by pixel-level Intersection over Union (IoU) and F1. Authentic predictions yield all-zero masks.

\begin{sloppypar}
We fine-tune Qwen3-VL-8B-Instruct~\cite{qwen3vl} with LoRA ($r{=}128$, $\alpha{=}256$) on all linear layers; the vision encoder is frozen. SFT: 3 epochs, 8$\times$A100-80GB, DeepSpeed ZeRO-3~\cite{rajbhandari2020zero}, lr $2{\times}10^{-5}$, cosine schedule, warmup 0.1, batch 32, max 32768 tokens, loss on assistant turns only.
RL: GRPO ($G{=}8$, $\beta_{\text{KL}}{=}0.001$, $\epsilon^{-}{=}0.2$, $\epsilon^{+}{=}0.3$, $\tau{=}0.6$, top-$p{=}0.90$, $\beta_{\text{ent}}{=}0.0002$), batch 128, 400 steps. Curriculum decay: $t_0$ at 50\% of total steps, $\alpha{=}5.0$. Reward weights: $w_1{=}0.35$, $w_2{=}0.35$, $w_3{=}0.25$, $w_4{=}0.05$. Qwen3-VL-235B-A22B~\cite{qwen3vl} serves as the MLLM judge.

We compare: (1)~Conventional IMDL detectors (MVSS-Net~\cite{mvssnet}, TruFor~\cite{trufor}, ForMa~\cite{forma}, SparseViT~\cite{sparsevit})---pixel masks without explanations; (2)~MLLM-based detectors (FakeShield~\cite{fakeshield}, SIDA-7B~\cite{sida})---interpretable but lacking forensic grounding; (3)~General-purpose MLLMs (GPT-5.4~\cite{singh2025openai}, Gemini-2.5-Pro~\cite{comanici2025gemini}, Qwen3-VL-8B~\cite{qwen3vl})---zero-shot without task-specific training; plus domain-specific baselines: APSC-Net~\cite{miml} for documents, UnivFD~\cite{univfd}, DRCT~\cite{drct} for AIGC. (4)~Ablation variants in \S\ref{sec:ablation}.
\end{sloppypar}

% ====================================================================
% Table 2: Document Forgery Detection (single-column)
% ====================================================================
\begin{table}[ht]
\centering
\caption{Document manipulation detection and localization on T-SROIE and FSTS-1.5k (zero-shot, localization only). \textbf{Bold} = best, \underline{underline} = second.}
\label{tab:doc}
\renewcommand{\arraystretch}{0.94}
\setlength{\tabcolsep}{4.8pt}
\small
\begin{tabular}{l cc cc cc}
\toprule
 & \multicolumn{2}{c}{T-SROIE}
 & \multicolumn{2}{c}{FSTS-1.5k}
 & \multicolumn{2}{c}{\textbf{Avg}} \\
\cmidrule(lr){2-3} \cmidrule(lr){4-5} \cmidrule(lr){6-7}
\multicolumn{1}{c}{Method}
 & pF1 & IoU
 & pF1 & IoU
 & pF1 & IoU \\
\midrule
% --- IFDL methods ---
MVSS-Net~\cite{mvssnet} (\textit{ICCV'21})
 & .024 & .012 & .124 & .077 & .074 & .045 \\
TruFor~\cite{trufor} (\textit{CVPR'23})
 & .061 & .037 & .406 & .317 & .234 & .177 \\
PSCC-Net~\cite{psccnet} (\textit{TCSVT'22})
 & .019 & .009 & .346 & .264 & .183 & .137 \\
APSC-Net~\cite{miml} (\textit{CVPR'24})
 & .186 & .126 & .231 & .169 & .209 & .148 \\
ForMa (\textit{SPL'25})
 & .037 & .021 & .189 & .135 & .113 & .078 \\
SparseViT (\textit{AAAI'25})
 & .019 & .009 & .104 & .058 & .062 & .034 \\
\midrule
% --- MLLM-based IFDL methods ---
FakeShield~\cite{fakeshield} (\textit{ICLR'25})
 & .026 & .013 & .093 & .053 & .060 & .033 \\
SIDA-7B~\cite{sida} (\textit{CVPR'25})
 & .000 & .000 & .007 & .007 & .004 & .004 \\
\midrule
% --- General-purpose MLLMs ---
GPT-5.4 & .027 & .014 & .118 & .071 & .073 & .043 \\
Gemini-2.5-Pro & .022 & .012 & .042 & .024 & .032 & .018 \\
Qwen3-VL-8B & .026 & .013 & .122 & .073 & .074 & .043 \\
\midrule
% --- Ours ---
\rowcolor{gray!10} ATAR (sft only)
 & \underline{.203} & \underline{.132} & \underline{.505} & \underline{.434} & \underline{.354} & \underline{.283} \\
\rowcolor{gray!10} \textbf{ATAR (Ours)}
 & \textbf{.210} & \textbf{.140} & \textbf{.520} & \textbf{.450} & \textbf{.365} & \textbf{.295} \\
\bottomrule
\end{tabular}
\end{table}

\subsection{Main Results}
\label{sec:main_results}

We evaluate ATAR on image-level classification and pixel-level localization across all test sets, covering IMDL (Table~\ref{tab:imdl}), DMDL (Table~\ref{tab:doc}), Deepfake detection (Table~\ref{tab:deepfake}), and AIGC detection (Table~\ref{tab:aigc}).

\textbf{IMDL.}\quad On the six zero-shot IMDL benchmarks (Table~\ref{tab:imdl}), ATAR ranks first in average iF1 (78.5\%), ACC (74.6\%), pF1 (56.4\%), and IoU (49.6\%), surpassing TruFor by +5.6 percentage points (pp) in iF1 and +6.0\,pp in pF1. SFT-only already outperforms TruFor (74.3\% vs.\ 72.9\%); GRPO adds +4.2\,pp. Recent lightweight detectors ForMa~\cite{forma} and SparseViT~\cite{sparsevit} achieve competitive per-dataset localization on in-distribution data (ForMa: 62.9\% pF1 on CASIA v1+, 48.7\% on Coverage) but suffer severe cross-dataset degradation (ForMa drops to 0.1\% on Columbia and 5.5\% on NIST16; SparseViT to 0.0\% on Columbia and 15.3\% on CASIA v1+), and their image-level detection remains near chance (iF1 $\approx$ 66--69\%) due to the lack of a dedicated classification head. Among MLLMs, FakeShield trails by 11.8\,pp iF1; Qwen3-VL-8B---ATAR's base model---reaches only 29.0\% ($-$49.5\,pp), as visual encoders cannot perceive signal-level traces that ATAR externalizes to 22 forensic tools.

\textbf{DMDL.}\quad For DMDL (Table~\ref{tab:doc}), ATAR averages 36.5\% pF1, the best among all methods and far above TruFor (23.4\%) and GPT-5.4 (7.3\%). Conventional detectors collapse on documents (all below 23.4\% pF1) because document images contain dense, structured text with uniform backgrounds, where pixel-level artifacts from character-level tampering are far subtler than those in natural-image splicing; ATAR's tool-augmented grounding enables region-level inspection that isolates these fine-grained inconsistencies.

% ====================================================================
% Table 3: Deepfake Detection (single-column)
% ====================================================================
\begin{table}[ht]
\centering
\small
\caption{Deepfake detection on OpenForensics (in-domain). \textbf{Bold} = best, \underline{underline} = second.}
\label{tab:deepfake}
\renewcommand{\arraystretch}{0.94}
\setlength{\tabcolsep}{10pt}
\begin{tabular}{l cccc}
\toprule
\multicolumn{1}{c}{Method} & iF1 & ACC & pF1 & IoU \\
\midrule
% --- IFDL methods ---
TruFor~\cite{trufor} (\textit{CVPR'23})
 & .999 & .998 & .423 & .320 \\
ForMa (\textit{SPL'25})
 & .999 & .998 & .369 & .272 \\
SparseViT (\textit{AAAI'25})
 & .999 & .998 & .103 & .058 \\
\midrule
% --- MLLM-based IFDL methods ---
FakeShield~\cite{fakeshield} (\textit{ICLR'25})
 & .433 & .284 & .122 & .073 \\
SIDA-7B~\cite{sida} (\textit{CVPR'25})
 & .015 & .009 & .006 & .005 \\
\midrule
% --- General-purpose MLLMs ---
GPT-5.4 & .113 & .370 & .098 & .061 \\
Gemini-2.5-Pro & .293 & .420 & .092 & .056 \\
Qwen3-VL-8B & .034 & .320 & .112 & .066 \\
\midrule
% --- Ours ---
\rowcolor{gray!10} ATAR (sft only)
 & .999 & .998 & \underline{.566} & \underline{.431} \\
\rowcolor{gray!10} \textbf{ATAR (Ours)}
 & .999 & .998 & \textbf{.616} & \textbf{.520} \\
\bottomrule
\end{tabular}
\end{table}

\textbf{Deepfake.}\quad For Deepfake detection (Table~\ref{tab:deepfake}), ATAR reaches 99.9\% iF1 and 61.6\% pF1, matching TruFor in classification while outperforming it in localization by +19.3\,pp pF1. General-purpose MLLMs fail entirely (GPT-5.4: 11.3\%; Qwen3-VL-8B: 3.4\%).

% ====================================================================
% Table 4: AIGC Detection (single-column)
% ====================================================================
\begin{table}[ht]
\centering
\caption{AIGC detection on GenImage++ (zero-shot, classification only). \textbf{Bold} = best, \underline{underline} = second.}
\label{tab:aigc}
\renewcommand{\arraystretch}{0.94}
\setlength{\tabcolsep}{2pt}
\resizebox{\columnwidth}{!}{%
\scriptsize
\begin{tabular}{l cc cc cc cc cc cc}
\toprule
 & \multicolumn{2}{c}{SD 1.5}
 & \multicolumn{2}{c}{SDXL}
 & \multicolumn{2}{c}{SD 3}
 & \multicolumn{2}{c}{SD3-R}
 & \multicolumn{2}{c}{Flux}
 & \multicolumn{2}{c}{Avg.} \\
\cmidrule(lr){2-3} \cmidrule(lr){4-5} \cmidrule(lr){6-7} \cmidrule(lr){8-9} \cmidrule(lr){10-11} \cmidrule(lr){12-13}
\multicolumn{1}{c}{Method}
 & F1 & ACC & F1 & ACC & F1 & ACC & F1 & ACC & F1 & ACC & F1 & ACC \\
\midrule
% --- AIGC-specific detection methods ---
UnivFD~\cite{univfd} (\textit{CVPR'23})
 & .218 & .146 & .233 & .155 & .044 & .099 & .024 & .089 & .009 & .082 & .106 & .114 \\
DRCT~\cite{drct} (\textit{ICML'24})
 & \underline{.927} & \underline{.867} & .473 & .327 & \underline{.722} & .598 & .504 & .388 & .501 & .385 & .625 & .513 \\
ForMa~\cite{forma} (\textit{SPL'25})
 & .653 & .485 & .667 & .500 & .667 & .500 & .667 & .500 & .667 & .500 & .664 & .497 \\
SparseViT~\cite{sparsevit} (\textit{AAAI'25})
 & .667 & .500 & .667 & .500 & .667 & .500 & .667 & .500 & .667 & .500 & .667 & .500 \\
\midrule
% --- MLLM-based IFDL methods ---
FakeShield~\cite{fakeshield} (\textit{ICLR'25})
 & .667 & .505 & .660 & .495 & .596 & .425 & .625 & .455 & .556 & .385 & .621 & .453 \\
SIDA-7B~\cite{sida} (\textit{CVPR'25})
 & .157 & .515 & .514 & .660 & .547 & \underline{.660} & \underline{.878} & \underline{.880} & \textbf{.881} & \textbf{.885} & .595 & .720 \\
\midrule
% --- General-purpose MLLMs ---
GPT-5.4
 & .574 & .420 & .783 & .650 & .255 & .240 & .651 & .700 & .487 & .390 & .550 & .480 \\
Gemini-2.5-Pro
 & \textbf{.940} & \textbf{.890} & \textbf{.974} & \textbf{.950} & .705 & .590 & .762 & .810 & \underline{.841} & .750 & \underline{.844} & \underline{.798} \\
Qwen3-VL-8B
 & .530 & .380 & .444 & .300 & .576 & .470 & .828 & .712 & .784 & .680 & .632 & .508 \\
\midrule
% --- Ours ---
\rowcolor{gray!10} \textbf{ATAR (Ours)}
 & .925 & .862 & \underline{.885} & \underline{.860} & \textbf{.830} & \textbf{.728} & \textbf{.882} & \textbf{.896} & .828 & \underline{.766} & \textbf{.870} & \textbf{.822} \\
\bottomrule
\end{tabular}%
}
\end{table}

\textbf{AIGC.}\quad For AIGC detection (Table~\ref{tab:aigc}), ATAR is never trained on AIGC data, yet achieves 87.0\% average F1 across five generators, outperforming all general-purpose MLLMs (Gemini-2.5-Pro: 84.4\%) and AIGC-specific methods such as UnivFD and DRCT.

% ====================================================================
% Table 5: Unified Ablation Study (single-column)
% ====================================================================
\subsection{Ablation Study}
\label{sec:ablation}

We ablate ATAR along three dimensions on the IMDL benchmark (6-dataset zero-shot average). Results are reported in Table~\ref{tab:ablation}.

\textbf{Tool Augmentation.}\quad The ordering Full (.785) $\gg$ w/o Tools (.751) $>$ Random Tool (.726) shows that the value of tools lies in selecting the right one: a random heatmap is worse than no heatmap ($-$2.5\,pp), because the model treats irrelevant highlights as evidence and commits to wrong verdicts with high confidence.

\textbf{Training Pipeline.}\quad SFT teaches the multi-turn format but not tool selection (ToolAcc 41.8\%); RL raises it to 72.0\%. Without the tool prior curriculum, RL degrades ToolAcc to 38.6\%---below SFT-only's 41.8\%---as failed random attempts produce negative gradients that collapse selection to a few broad-spectrum tools. Fixed curriculum (no decay) recovers ToolAcc to 63.4\% but the model never internalizes autonomous selection (iF1 $-$1.2\,pp vs.\ Full). Only progressive decay yields both high ToolAcc and generalization.

\textbf{Reward Design.}\quad Each reward component prevents a distinct failure mode. Without $r_2$, the model over-predicts forgery (ACC drops 4.8\,pp) since it can maximize $r_1$ without precise localization. Without $r_3$, quantitative metrics barely change ($-$0.6\,pp iF1) but reasoning degenerates to shortcut pattern-matching---the value of $r_3$ is explainability, not accuracy ---cf.\ Table~\ref{tab:reasoning_quality}, where TP hallucination triples from 7.2\% to 21.6\%. Holistic Reward shows non-uniform degradation (iF1 $-$2.1\,pp vs.\ pF1 $-$6.2\,pp vs.\ ToolAcc $-$12.2\,pp): a single score is dominated by classification, starving localization and tool selection of gradient signal.

\begin{table}[ht]
\centering
\caption{Ablation study (IMDL 6-dataset zero-shot average). ToolAcc\,=\,Top-$K$ ($K{=}3$) optimal tool hit rate. \textbf{Bold}\,=\,best.}
\label{tab:ablation}
\renewcommand{\arraystretch}{0.88}
\setlength{\tabcolsep}{3pt}
\resizebox{\columnwidth}{!}{%
\footnotesize
\begin{tabular}{l cccc c}
\toprule
\multicolumn{1}{c}{Variant}
 & iF1 & ACC & pF1 & IoU & ToolAcc \\
\midrule
\rowcolor{gray!10} \textbf{ATAR (Full)}
 & \textbf{.785} & \textbf{.746} & \textbf{.564} & \textbf{.496} & \textbf{.720} \\
\midrule
\multicolumn{6}{l}{\textit{(a) Tool Augmentation}} \\
\quad w/o Tools
 & .751 & .714 & .476 & .410 & -- \\
\quad Random Tool
 & .726 & .694 & .426 & .364 & .045 \\
\midrule
\multicolumn{6}{l}{\textit{(b) Training Pipeline}} \\
\quad SFT-only (no RL)
 & .743 & .711 & .468 & .402 & .418 \\
\quad w/o Curriculum
 & .756 & .716 & .494 & .426 & .386 \\
\quad Fixed Curriculum (no decay)
 & .773 & .736 & .534 & .470 & .634 \\
\midrule
\multicolumn{6}{l}{\textit{(c) Reward Design}} \\
\quad w/o $r_3$ (reasoning quality)
 & .779 & .740 & .551 & .484 & .704 \\
\quad Holistic Reward
 & .764 & .724 & .502 & .438 & .598 \\
\quad w/o $r_2$ (localization)
 & .762 & .698 & .452 & .384 & .682 \\
\bottomrule
\end{tabular}%
}
\end{table}

% ====================================================================
% Qualitative Analysis
% ====================================================================
\subsection{Qualitative Analysis}
\label{sec:qualitative}

\subsubsection{\textbf{Reasoning Quality}}
\label{sec:reasoning_quality}
We evaluate explanation quality on 200 stratified-sampled images from the six zero-shot IMDL datasets. An MLLM judge (Qwen3-VL-235B-A22B) receives the original image, ground-truth label, and the method's anonymized explanation; for ATAR variants it additionally receives tool-generated heatmaps to verify faithfulness. Each dimension is scored on a 5-point Likert scale following LLM-as-judge practice~\cite{zheng2023judging}: \textbf{Groundedness}, \textbf{Consistency}, \textbf{Faithfulness}, and \textbf{Specificity}. We also report the Hallucination Rate---fraction of explanations with at least one fabricated claim---overall and on true positives (TP). A 50-sample human evaluation (3 annotators) validates the judge via Fleiss' $\kappa$ and Spearman $\rho$.

% ====================================================================
% Table: Reasoning Quality (merged)
% ====================================================================
\begin{table}[ht]
\centering
\caption{Reasoning quality (200 zero-shot IMDL samples, scored 1--5). Hallu.\,Rate\,=\,fraction with fabricated evidence. $\kappa$/$\rho$\,=\,human--judge agreement.}
\label{tab:reasoning_quality}
\renewcommand{\arraystretch}{0.88}
\setlength{\tabcolsep}{3pt}
\resizebox{\columnwidth}{!}{%
\footnotesize
\begin{tabular}{l cccc c cc}
\toprule
 & \multicolumn{5}{c}{Reasoning Quality (1--5)\,$\uparrow$}
 & \multicolumn{2}{c}{Hallu.\ Rate\,$\downarrow$} \\
\cmidrule(lr){2-6} \cmidrule(lr){7-8}
Method
 & Ground. & Consist. & Faith. & Specif. & Avg
 & All & TP \\
\midrule
FakeShield
 & 2.78 & 3.12 & 2.56 & 2.88 & 2.84
 & 38.5\% & 29.2\% \\
Gemini-2.5-Pro
 & 2.32 & 3.36 & 2.41 & 2.74 & 2.71
 & 44.0\% & 34.8\% \\
ATAR (w/o $r_3$)
 & 3.42 & 3.48 & 3.18 & 3.24 & 3.33
 & 27.5\% & 21.6\% \\
\rowcolor{gray!10} \textbf{ATAR}
 & \textbf{4.18} & \textbf{4.06} & \textbf{4.28} & \textbf{3.96} & \textbf{4.12}
 & \textbf{11.5\%} & \textbf{7.2\%} \\
\midrule
Fleiss' $\kappa$
 & .68 & .54 & .72 & .58 & --
 & -- & -- \\
Spearman $\rho$
 & .82 & .74 & .86 & .76 & --
 & -- & -- \\
\bottomrule
\end{tabular}%
}
\end{table}

ATAR leads all dimensions (Avg 4.12 vs.\ 3.33 runner-up), with the largest gains in Faithfulness (+1.10) and Groundedness (+0.76)---tool outputs provide verifiable anchors that prevent fabrication. Gemini-2.5-Pro shows a revealing split: highest non-ATAR Consistency (3.36) but lowest Groundedness (2.32) and Faithfulness (2.41), as it fabricates forensic terminology (e.g., ``ELA analysis reveals\ldots'') without any tool to run.

The most critical finding is the TP Hallucination Rate, which measures shortcut reasoning: correct verdict, fabricated justification. Removing $r_3$ triples TP hallucination from 7.2\% to 21.6\% while iF1 drops only 0.6\,pp per Table~\ref{tab:ablation}---classification metrics entirely mask this degradation. Gemini-2.5-Pro reaches 34.8\%, confirming that general-purpose MLLMs routinely fabricate evidence even on correct predictions. Human--Judge agreement ($\kappa$\,=\,.54--.72; $\rho$\,=\,.74--.86) validates the automated evaluation, with highest reliability on Faithfulness ($\rho{=}.86$).

\subsubsection{\textbf{Tool Specialization}}
\label{sec:tool_dynamics}
Table~\ref{tab:tool_selection} compares SFT and post-RL tool domain selection on the six zero-shot IMDL test sets.

% ====================================================================
% Table: Tool Domain Selection Distribution
% ====================================================================
\begin{table}[ht]
\centering
\caption{Tool domain selection (\%) on zero-shot IMDL sets: SFT vs.\ post-RL. \textbf{Bold}\,=\,Top-1 per dataset (RL). $\Delta$Hit\,=\,optimal-set hit rate.}
\vspace{-3mm}
\label{tab:tool_selection}
\renewcommand{\arraystretch}{0.86}
\setlength{\tabcolsep}{1.5pt}
\resizebox{\columnwidth}{!}{%
\scriptsize
\begin{tabular}{ll ccccccc c}
\toprule
 & & Cmpr. & Noise & Freq. & Str. & C-Mv. & AIGC & F.M. & $\Delta$Hit \\
\midrule
\multirow{2}{*}{CASIA v1+}
 & SFT & 32.4 & 20.6 & 10.8 & 9.2 & 6.8 & 12.4 & 7.8 & 62.8 \\
 & RL  & \textbf{46.3} & 24.8 & 5.2 & 6.4 & 4.2 & 7.8 & 5.3 & 82.4 \\
\multirow{2}{*}{CocoGlide}
 & SFT & 8.6 & 14.8 & 8.4 & 10.6 & 4.2 & 32.4 & 21.0 & 64.2 \\
 & RL  & 5.2 & 16.4 & 4.8 & 8.2 & 2.4 & \textbf{56.8} & 6.2 & 88.2 \\
\multirow{2}{*}{Coverage}
 & SFT & 6.4 & 16.2 & 8.6 & 12.4 & 18.4 & 22.8 & 15.2 & 58.6 \\
 & RL  & 3.1 & 12.6 & 4.2 & 8.4 & \textbf{42.8} & 18.4 & 10.5 & 91.6 \\
\multirow{2}{*}{Korus}
 & SFT & 18.2 & 22.6 & 14.8 & 10.4 & 5.6 & 16.8 & 11.6 & 56.4 \\
 & RL  & 22.6 & \textbf{32.8} & 10.4 & 6.2 & 3.2 & 14.2 & 10.6 & 76.8 \\
\multirow{2}{*}{Columbia}
 & SFT & 12.8 & 28.4 & 10.6 & 11.2 & 4.8 & 18.6 & 13.6 & 60.8 \\
 & RL  & 4.8 & \textbf{42.6} & 8.2 & 9.6 & 2.6 & 22.6 & 9.6 & 85.4 \\
\multirow{2}{*}{NIST16}
 & SFT & 28.6 & 22.4 & 9.8 & 10.2 & 8.4 & 12.8 & 7.8 & 58.2 \\
 & RL  & \textbf{38.7} & 26.4 & 5.6 & 7.2 & 7.8 & 8.6 & 5.7 & 78.2 \\
\bottomrule
\end{tabular}%
}
\end{table}

\begin{sloppypar}
SFT produces a diffuse distribution (no domain exceeds 33\%); GRPO sharpens it into dataset-specific specialization whose Top-1 domain matches the dominant forgery type: Compression for JPEG-compressed CASIA v1+/NIST16, AIGC for GLIDE-inpainted CocoGlide, Copy-Move for Coverage, and Noise for spliced Columbia and Korus. The model also learns to suppress irrelevant tools---e.g., Columbia's Compression drops from 12.8\% to 4.8\% because its uncompressed TIF format renders JPEG ghost tools uninformative. $\Delta$Hit rises from 56--65\% (SFT) to 77--92\% (RL), highest on Coverage (91.6\%, single dominant cue) and lowest on Korus (76.8\%, mixed forgery requiring diverse tools). Freq.\ and Str.\ domains decline uniformly ($-$3--6\,pp), pruned as generically weaker alternatives. SFT teaches when to call a tool; RL teaches which one.
\end{sloppypar}

% ====================================================================
% Case Study
% ====================================================================
\subsubsection{\textbf{Case Study}}
\label{sec:case_study}

The complete traces in the supplementary material show that ATAR can revise a
tool-refuted hypothesis, combine semantic and signal-level evidence, and
accumulate evidence across multiple regions. These cases illustrate that ATAR
adapts its reasoning strategy to each input instead of following a fixed
detection pipeline.

%% file: Content/conclusion.tex
\section{Conclusion}
We presented ATAR, an agentic framework that operationalizes the judicial forensic workflow of ``experimental analysis -- logical reasoning -- scientific evidence'' for explainable image forgery detection.
Building on dual-stream evidence from semantic inspection and 22 forensic tools, the agent performs logical reasoning by interpreting observations, testing hypotheses, and abandoning tool-refuted directions across multiple turns, with Forensics Curriculum Learning raising tool accuracy from 41.8\% to 72.0\%.
The Structured Evidence Reward ensures that the scientific evidence is faithful rather than fabricated, reducing TP hallucination to 7.2\% versus 34.8\% in general-purpose MLLMs, while achieving state-of-the-art detection across IMDL, Deepfake, DMDL, and AIGC benchmarks.